\documentclass[a4paper]{article}

\usepackage{INTERSPEECH2021}

\usepackage[dvipsnames]{xcolor}
\definecolor{mycolor}{HTML}{FF6600}
\definecolor{mycolor2}{HTML}{1E77B4}

\usepackage[prependcaption,textsize=scriptsize]{todonotes}
\usepackage{microtype}
\newcommand{\tablesystem}[1]{{\eightpt \textsc{#1}}}
\definecolor{mygreen}{HTML}{2CA02C}
\definecolor{myblue}{HTML}{1E77B4}
\definecolor{myred}{HTML}{D62728}
\newcommand{\captionsep}{\vspace*{-5pt}}

\usepackage{hyperref}
\hypersetup{hidelinks}

\usepackage{amsmath}
\usepackage{mathtools}
\DeclarePairedDelimiter{\norm}{\lVert}{\rVert}

\usepackage{booktabs} 
\usepackage{tabularx}
\usepackage{multirow}
\usepackage{siunitx}
\newcommand{\mytable}{
	\centering
	\renewcommand{\arraystretch}{1.2}
}
\newcolumntype{C}{>{\centering\arraybackslash}X}
\newcolumntype{L}{>{\raggedright\arraybackslash}X}

\title{Multilingual transfer of acoustic word embeddings improves when training on languages related to the target zero-resource language}

\name{Christiaan Jacobs \qquad Herman Kamper}
\address{E\&E Engineering, Stellenbosch University} 
\email{20111703@sun.ac.za, kamperh@sun.ac.za}

\usepackage{cite}

\begin{document}
	
	\maketitle
	\begin{abstract}	
    Acoustic word embedding models map variable duration speech segments to fixed dimensional vectors, enabling efficient speech search and discovery. Previous work explored how embeddings can be obtained in zero-resource settings where no labelled data is available in the target language. The current best approach uses transfer learning: a single supervised multilingual model is trained using labelled data from multiple well-resourced languages and then applied to a target zero-resource language (without fine-tuning). However, it is still unclear how the specific choice of training languages affect downstream performance. Concretely, here we ask whether it is beneficial to use training languages related to the target. Using data from eleven languages spoken in Southern Africa, we experiment with adding data from different language families while controlling for the amount of data per language.     In word discrimination and query-by-example search evaluations, we show that training on languages from the same family gives large improvements. Through finer-grained analysis, we show that training on even just a single related language gives the largest gain. We also find that adding data from unrelated languages generally doesn't hurt performance.
	\end{abstract}
		
	\section{Introduction}
\label{sec:introduction}

Developing robust speech systems for zero-resource languages---where no transcribed speech resources are available for model training---remains a challenge.
Although full speech recognition is not possible in most zero-resource settings, researchers have proposed methods for applications such as speech search~\cite{levin+etal_icassp15,huang+etal_arxiv18,yuan+etal_interspeech18},
word discovery~\cite{park+glass_taslp08,jansen+vandurme_asru11,ondel+etal_interspeech19,rasanen+blandon_arxiv20}, and segmentation and clustering~\cite{kamper+etal_asru17,seshadri+rasanen_spl19,kreuk+etal_interspeech20}, making use of only %
unlabelled speech audio.
Many of these applications require speech segments of different lengths to be compared.
This is conventionally done using alignment (e.g.\ with dynamic time warping). 
But this can be slow and inaccurate. %

Acoustic word embedding (AWE) models map a %
variable duration speech segment %
to a fixed dimensional vector~\cite{levin+etal_asru13}.
The goal is to map instances of the same word type to similar vectors.
Segments can then be efficiently compared by calculating the distance %
in the embedding space.
Given the advantages AWEs have over %
alignment methods, several AWE models have %
been proposed~\cite{bengio+heigold_interspeech14,he+etal_iclr17,audhkhasi+etal_stsp17,wang+etal_icassp18,chen+etal_slt18,holzenberger+etal_interspeech18,chung+glass_interspeech18,haque+etal_icassp19,shi+etal_slt21,palaskar+etal_icassp19,settle+etal_icassp19,jung+etal_asru19}. Many of these are for the supervised setting, using labelled data to train a discriminative model. %

For the zero-resource setting, a number of unsupervised AWE approaches have also been explored, many relying on autoencoder-based neural models trained on unlabelled data in the target language~\cite{chung+etal_interspeech16,kamper+etal_icassp16,holzenberger+etal_interspeech18,kamper_icassp19}.
However, there still exists a large performance gap between these unsupervised models and %
their supervised counterparts~\cite{levin+etal_asru13,kamper_icassp19}.
A recent alternative for obtaining AWEs on a zero-resource language is to use multilingual transfer learning~\cite{ma+etal_arxiv20,kamper+etal_icassp20,kamper+etal_taslp21,hu+etal_slt21, hu+etal_interspeech20}. %
The goal is %
to have the benefits of supervised learning by training a model on labelled data from multiple well-resourced languages, but to then
 apply
the model to an unseen target zero-resource language without fine-tuning it---a form of \textit{transductive} transfer learning~\cite{ruder_phd19}.
This multilingual transfer approach has been shown to outperform unsupervised monolingual AWE models~\cite{kamper+etal_taslp21,jacobs+etal_slt21}. %

Although %
there is clear benefit in applying multilingual AWE models
to an unseen zero-resource language, it is still unclear how the particular choice of training languages affects subsequent performance.
Preliminary experiments~\cite{kamper+etal_taslp21} show improved scores
when training a monolingual model on one language and applying it to another from the same family.
But this has not been investigated systematically and there are still several unanswered questions:
Does the benefit of training on related languages diminish as we train on more languages (which might or might not come from the same family as the target zero-resource language)?
When training exclusively on related languages, does performance suffer when adding an unrelated language?
Should we prioritise data set size or language diversity when collecting data for multilingual AWE transfer?

We try to answer these questions using %
a corpus of under-resourced languages spoken in Southern Africa.
These languages can be grouped into different families %
based on their linguistic links. %
We specifically
want to see whether it is beneficial to use closely related languages when training 
a multilingual model catered to a specific zero-resource language, similar to~\cite{yi+etal_taslp19}.
We divide the corpus into training and test languages, with (some of) the test languages coming from families that also occur in %
training. %
We conduct several experiments where we add data from different language families, and also %
control for the amount of data per language.
AWEs are evaluated
in an isolated word discrimination task and in query-by-example (QbE) speech search on full utterances.
To our knowledge, only one other %
study~\cite{hu+etal_slt21} has done
AWE-based QbE
using multilingual transfer.

Our main findings are as follows. 
(i)~Training a multilingual model using languages that are closely related to the target %
improves %
performance. %
This is true, not just
because of the increase in data, but because of %
language diversity. 
(ii)~When %
we systematically add training languages, the largest improvements are %
gained from adding a single related language. Adding more related languages %
gives small gains. 
Adding unrelated languages generally gives small or no %
gains, but also doesn't hurt.
(iii)~For a %
target language, the %
performance of a multilingual model trained on unrelated languages can be matched by a model trained with much less data from multiple related~languages.

	\section{Languages of South Africa}
\label{sec:african}

\begin{figure}[t]
    \centering
    \includegraphics[width=0.99\linewidth]{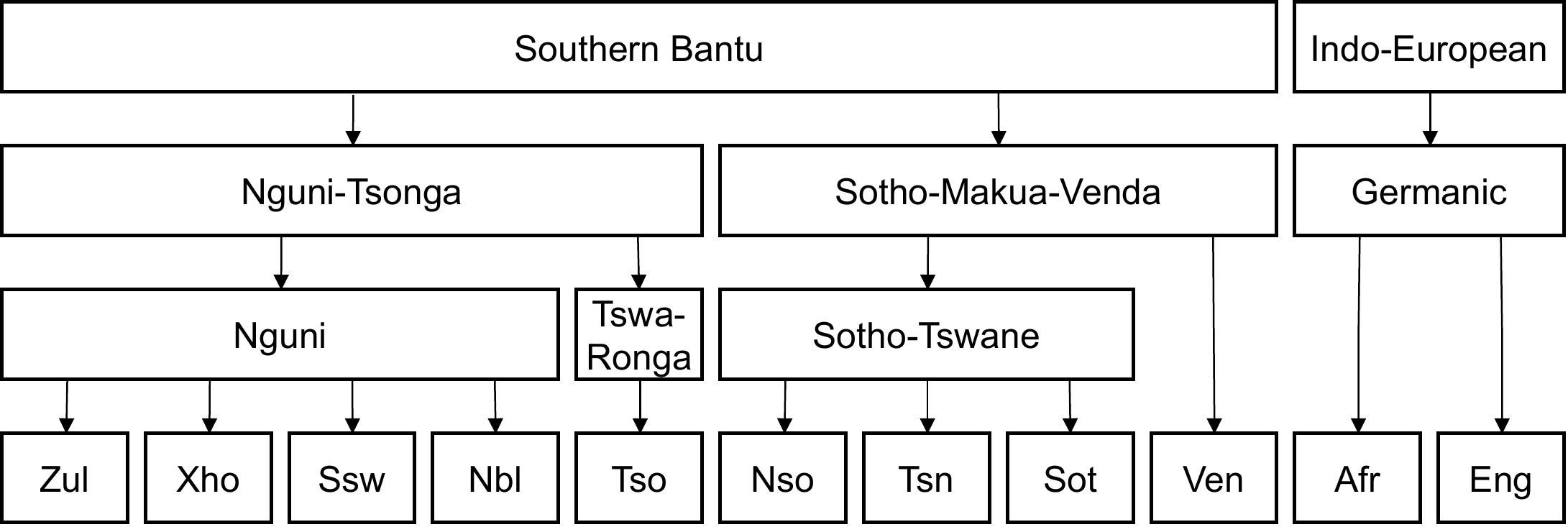}
    \captionsep
    \caption[Caption]{A family tree for the official South African languages.\footnotemark
    }
    
    \label{fig:hierarchy}
    \vspace{-5mm}
\end{figure}

The majority of languages in Africa are considered under-resourced~\cite{orife+etal_ICLR20}. %
This includes the eleven 
official languages of South Africa.
As show in Figure~\ref{fig:hierarchy}, nine of these languages %
belong to the larger Southern Bantu family: isiZulu (Zul), isiXhosa (Xho), Sepedi (Nso), Setswana (Tsn), Sesotho (Sot), Xitsonga (Tso), siSwati (Ssw), Tshivenda (Ven) %
and isiNdebele (Nbl). 
Many of these languages are also spoken %
in countries neighbouring South Africa.
From the Southern Bantu family there exist two principal, %
Nguni-Tsonga and Sotho-Makua-Venda. %
In the Sotho-Makua-Venda subfamily, Ven is somewhat of a standalone.
The other two languages, Afrikaans (Afr) and English (Eng), are Germanic languages from the %
Indo-European %
family.
All of these languages are considered under-resourced, except for Eng~\cite{barnard+etal_sltu14}.
To give an intuitive idea of how related these languages are, most of the Bantu languages  that are grouped together at the lowest level of the hierarchy %
would (to an %
extent) be intelligible to a native speaker of another languages in the same group.

\footnotetext{\scriptsize{\url{https://southafrica-info.com/arts-culture/11-languages-south-africa/}}}

	\section{Acoustic word embedding model}
\label{sec:model}

\begin{figure}[t]
	\centering
	\includegraphics[width=0.99\linewidth]{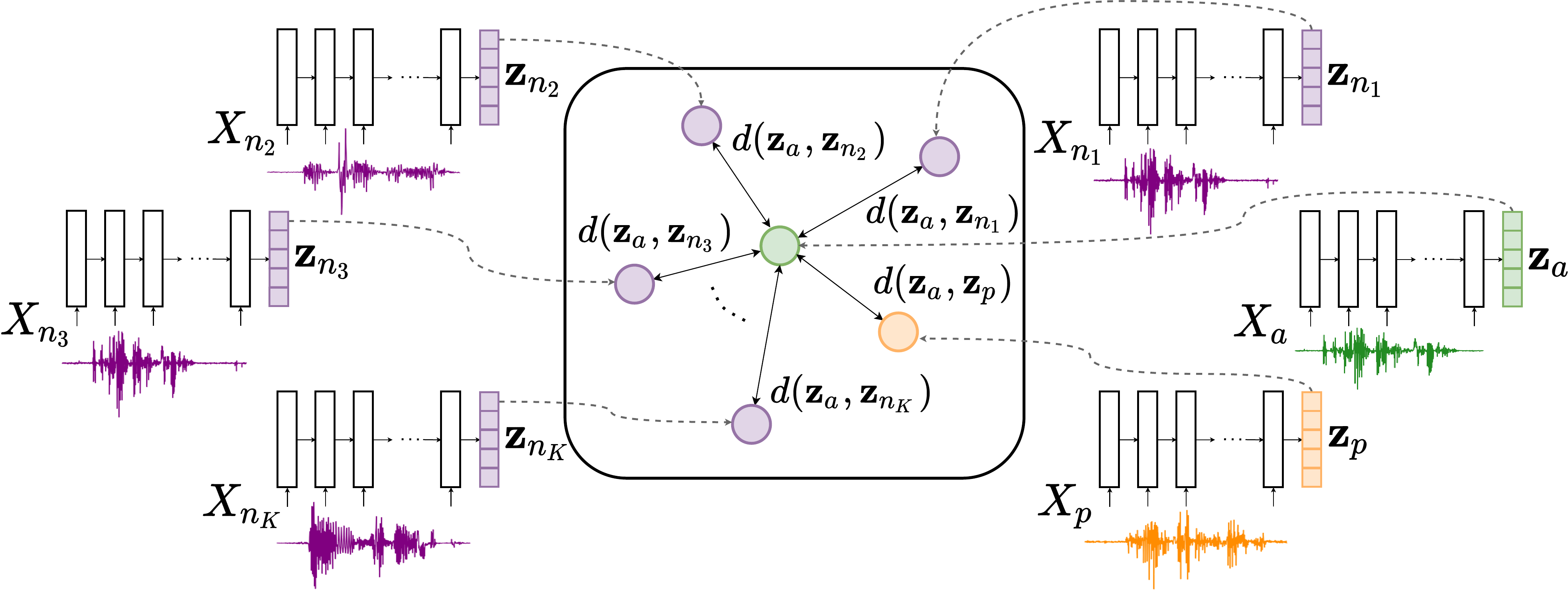}
    \captionsep
	\caption{The \tablesystem{ContrastiveRNN}.
	The model is trained to minimise the distance between the anchor and positive item $d(\mathbf{z}_a, \mathbf{z}_p)$ while maximising the distance between the anchor and multiple negatives $(\mathbf{z}_a, \mathbf{z}_{n_{k}})$.}
	\label{fig:contrastiveRNN}
	\vspace{-5mm}
\end{figure}

We use the \tablesystem{ContrastiveRNN} AWE model of~\cite{jacobs+etal_slt21}.\footnote{We extend the code available at \scriptsize{\url{https://github.com/christiaanjacobs/globalphone_awe_pytorch}}.}
It performed the best of the model variants considered for multilingual transfer in~\cite{jacobs+etal_slt21}. %
The model consists of an encoder recurrent neural network (RNN) %
that produces fixed-dimensional representations from variable-length speech segments. 
It is trained %
to minimise %
the distance between embeddings from %
speech segments %
of the same word type while %
maximising the distance between embeddings from multiple words of a different type. 
Let's use $X = \mathbf{x}_1, \mathbf{x}_2, \ldots, \mathbf{x}_T$ to denote a sequence of speech features.
Formally, given %
speech segments
$X_a$ and $X_p$ containing instances of the same word type and multiple negative examples $X_{n_{1}}, \ldots, X_{n_{K}}$, the \tablesystem{ContrastiveRNN} produces embeddings $\mathbf{z}_a, \mathbf{z}_p, \mathbf{z}_{n_{1}}, \ldots, \mathbf{z}_{n_{K}}$
(subscripts indicate anchor, positive and negative, respectively).
Each embedding is a fixed dimensional vector $\mathbf{z} \in \mathbb{R}^M$.
The model is illustrated in Figure~\ref{fig:contrastiveRNN}.
Let $\text{sim}(\mathbf{u}, \mathbf{v}) = \mathbf{u}^{\top}\mathbf{v}/\norm{\mathbf{u}}\norm{\mathbf{v}}$ denote the cosine similarity between vectors $\mathbf{u}$ and $\mathbf{v}$.
The loss given a positive pair $(X_a, X_p)$ and the set of negative examples is then defined as~\cite{chen+etal_icml20}:
\vspace*{-2.5pt}
\begin{equation}
J = -\text{log}\frac{\text{exp}\big\{\text{sim}(\mathbf{z}_a, \mathbf{z}_p)/\tau\big\}}{\sum_{j \in \{p, n_1, \hdots, n_K\}}^{}\text{exp}\big\{\text{sim}(\mathbf{z}_a, \mathbf{z}_j)/\tau\big\}}\,\text{,}
\label{eqn:contrastive_loss}
\end{equation} 
where $\tau$ is a temperature parameter, tuned on development data.

In the zero-resource setting we don't have labelled data in the target {language} %
to construct the positive and negative %
word pairs required for training.
We therefore follow the approach of~\cite{kamper+etal_taslp21}, and train a multilingual model on ground truth word pairs (extracted from forced alignments) from a number of %
languages for which we have labelled data. 
Subsequently, at test time, we apply the encoder RNN from the multilingual model to extract AWEs for speech from
the target zero-resource language.

	\section{Query-by-example speech search} %
\label{sec:qbe}

For evaluating the different AWE {models} 
we use %
an isolated word discrimination task~\cite{carlin+etal_icassp11}.
Recent findings~\cite{algayres+etal_interspeech20} suggest, however, that this evaluation is not always indicative of downstream system performance.
We therefore also perform query-by-example (QbE) speech search, which in contrast to %
{the word discrimination task}, does not assume a test set of isolated words, but instead operates on full unsegmented utterances.

Concretely, QbE speech search is the task of identifying the utterances in a speech collection that contain instances of a given spoken query.
A number of approaches have been put forward for AWE-based QbE~\cite{levin+etal_icassp15, chen+etal_icassp15, settle+etal_interspeech17}. %
Here we use the simplified approach from~\cite{kamper+etal_icassp19}. %
Using an AWE model, we first embed the query segment.
If we knew the word boundaries in the search collection, we could embed each of the words in an utterance and simply look up the closest embeddings to the query.
Instead, because we do not have word boundaries, each utterance is split into overlapping segments from some minimum to some maximum duration.
Each segment from each utterance is then embedded separately using the AWE model.
Finally, to do the QbE task, the query embedding is compared to each of the utterance sub-segment embeddings (using cosine distance), and the minimum distance over the utterance is then taken as the score for whether the utterance contains the given query. %

	\section{Experimental setup}
\label{sec:experiments}

\textbf{Data.} We perform all our experiments on the NCHLT speech corpus~\cite{barnard+etal_sltu14}, which provides wide-band speech from each of the eleven official South African languages (\S\ref{sec:african}). %
We use a version of the corpus  
where all repeated utterances were removed, leaving %
roughly 56 hours of speech from around 200 speakers in each language.
We use the default training, validation and test sets. %
We treat six of the languages as well-resourced: %
Xho$^*$, Ssw$^*$, Nbl$^*$, Nso$^\dagger$, Tsn$^\dagger$ and Eng$^\ddagger$. 
We use labelled data from the well-resourced languages to train a single supervised multilingual model and then apply the model to the target zero-resource languages.
More specifically, %
we extract 100k true positive word pairs using forced alignments for each training language.
We select Zul$^*$, Sot$^\dagger$, Afr$^\ddagger$, and Tso as our zero-resource languages and use another language, Ven, for validation of each model.\footnote{We use superscripts to indicate the different language families: $^*$Nguni, $^\dagger$Sotho-Tswana, $^\ddagger$Germanic.} %
Training and evaluation languages are %
carefully selected such that for each evaluation language at least one language from the same %
family is part of the training languages, except for Tso which is in a group of its own.

\label{subsec:awe_models}

\textbf{Models.}
All speech audio is parametrised as  %
13-dimensional
static Mel-frequency cepstral coefficients. 
The encoder unit of the \tablesystem{ContrastiveRNN} models (\S\ref{sec:model}) consists of three unidirectional RNNs with 400-dimensional hidden vectors, with an embedding size of $M = 130$ %
dimensions.
Models are optimised using Adam optimisation~\cite{kingma+ba_iclr15} with a learning rate of $0.001$. 
The temperature parameter $\tau$ in~\eqref{eqn:contrastive_loss} is set to 0.1.

\textbf{Evaluation.}
We consider two tasks for evaluating the performance of the AWE models. First, we
use the \textit{same-different} %
word discrimination task~\cite{carlin+etal_icassp11} to measure the intrinsic quality of the %
AWEs.
To evaluate a particular AWE model, a set of isolated test word segments is embedded.
We use roughly 7k isolated word instances per language from the test data. %
For every word pair in this set, the cosine distance between their embeddings is calculated.
Two words can then be classified as being of the same or different type based on some distance threshold, and a precision-recall curve is obtained by varying the threshold.
The area under this curve is used as final evaluation metric, referred to as the average precision (AP).
We are particularly interested in obtaining embeddings that are speaker invariant.
As in~\cite{hermann+etal_csl21}, %
we therefore calculate AP by only taking the recall over instances of the same word spoken by different speakers. %

The second task is QbE (\S\ref{sec:qbe}). %
For each evaluation language we use approximately two hours of test %
utterances as the search %
collection.
Sub-segments for the utterances in the speech collection are obtained by embedding
windows stretching from 20 to 60 frames with a 3-frame overlap. %
For each evaluation language we randomly %
draw instances of 15 spoken query word types from a disjoint speech set (the development set---which we never use for any validation experiments) where we only consider query words with 
at least 5 characters
for Afr and Zul and 3 for Sot.
There are
between 6 and 51 occurrences of each query word.
For each QbE test, we ensure that the relevant
multilingual AWE model %
has not seen any of the search or query data
during training or validation.
We report precision at ten
($P@10$), %
which is the fraction of the ten top-scoring retrieved utterances from the search collection that %
 contains the given query.

	\section{Experimental results}
\label{sec:results}

\begin{figure}[!t]
	\centering
	\includegraphics[width=0.99\linewidth]{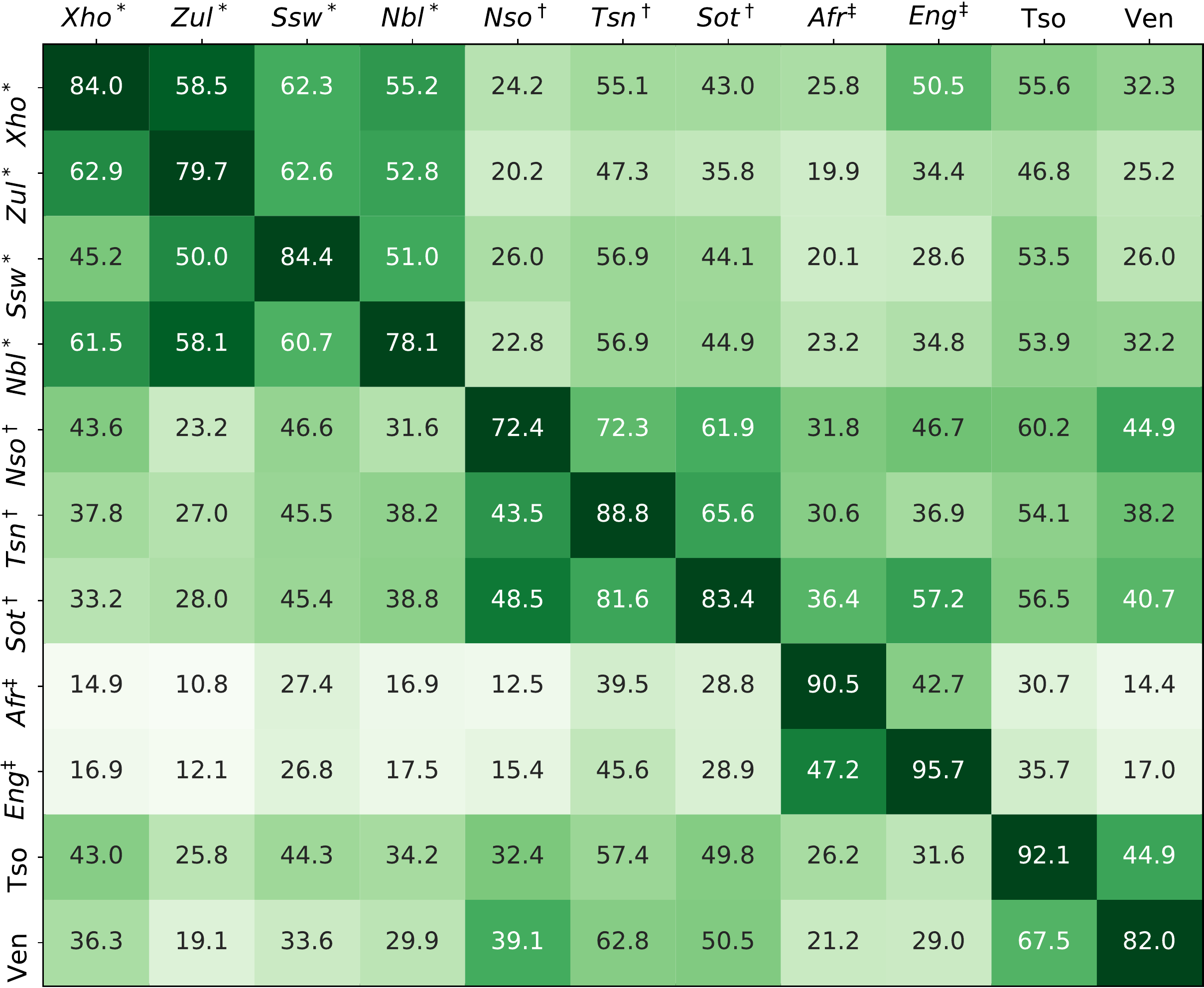}
    \captionsep
	\caption{AP (\%) 
    when training a monolingual supervised \tablesystem{ContrastiveRNN} on each language (rows) and then evaluating it on each of the other languages (columns). Heatmap colours are normalised for each evaluation language (i.e. per column).}
	\label{fig:heatmap}
	\vspace{-4mm}
\end{figure}

\subsection{Cross-lingual evaluation}

Before looking at multilingual modelling, we first consider a cross-lingual evaluation where we treat each language as a training language, train a supervised monolingual AWE model, and then apply it to every other language.
This allows us to see the effect of training on related languages in a pairwise fashion.
The results are shown in Figure~\ref{fig:heatmap}.
For each evaluation language excluding Tso and Ven, which are in family groups of their own,  %
the best results are achieved from models trained on a language from the same family.
E.g.\ on Zul, Xho is the best training language giving an AP of 58.5\%.
Eng is the only exception where the model trained on Sot performs better than using Afr, the other Germanic language.
Although Ven is in its own group at the lowest layer of the family tree in Figure~\ref{fig:hierarchy}, some of the best results when evaluating on Ven are obtained using models trained on Sotho-Tswana languages (Sot, Tsn, Nso), which are in the same family at a higher level.
We also see that for all nine Bantu evaluation languages, worst performance is obtained from the two Germanic models (Afr, Eng).

\begin{table}[!t]
	\mytable
	\caption{AP (\%) on test data for multilingual models trained on different combinations of well-resourced languages. Models are applied to two zero-resource languages from different language families, Nguni and Sotho-Tswana. For each training language 100k word pairs were extracted.}
    \captionsep
	\vspace{-1mm}    
    \eightpt
	\begin{tabularx}{1\linewidth}{Lcc}
		\toprule
		Multilingual model & Zul$^*$ & Sot$^\dagger$ \\
		\midrule
		\underline{\textit{Nguni:}} & &\\[2pt]        
		Xho$^*$ + Ssw$^*$ + Nbl$^*$ &\textbf{68.6} &---  \\
		Xho$^*$ + Ssw$^*$ + Eng$^\ddagger$ &60.9 &---  \\
		Xho$^*$ + Nso$^\dagger$ + Eng$^\ddagger$ &55.7 &---  \\
		Tsn$^*$ + Nso$^\dagger$ + Eng$^\ddagger$ &37.5 &---  \\
		Xho$^*$ + Ssw$^*$ + Nbl$^*$ {(\scriptsize{$10$\% subset})} &58.6 &---  \\[2pt]
		\underline{\textit{Sotho-Tswana:}} \\[2pt]
		Nso$^\dagger$ + Tsn$^\dagger $&--- &\textbf{76.7} \\
		Nso$^\dagger$ + Eng$^\ddagger$ &--- &64.8 \\		
		Xho$^*$ + Ssw$^*$ &--- &51.9 \\
		Xho$^*$ + Eng$^\ddagger$ &--- &52.5 \\		 
		Nso$^\dagger$ + Tsn$^\dagger$ {(\scriptsize{$10$\% subset})} &--- &58.4 \\       
		\bottomrule		
	\end{tabularx}
	\label{tbl:multilingual}
	\vspace{-4mm}
\end{table}

\begin{figure*}[t]
	\centering
	\includegraphics[scale=0.31]{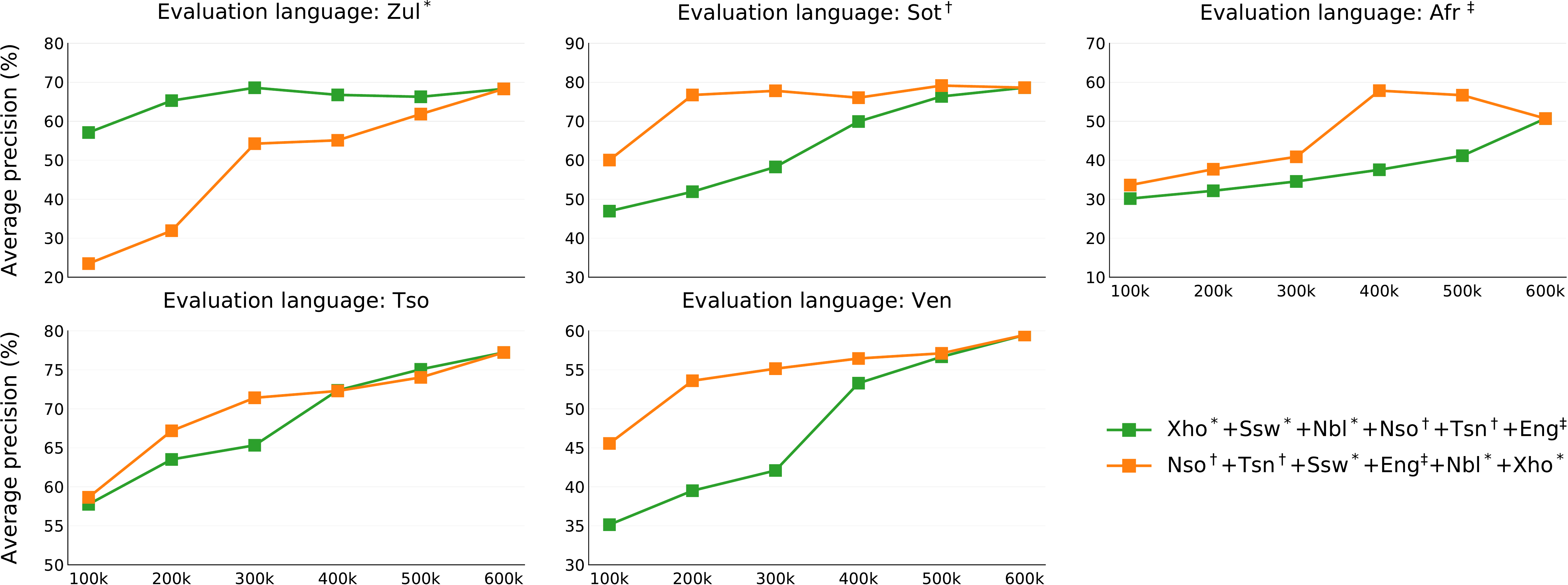}    
    \captionsep
	\caption{Same-different results %
    from two sequences of multilingual models, trained by adding one language at a time. For each training language, 100k positive word pairs are used, which is indicated on the $x$-axis.} %
	\label{fig:multi_increment}

    \vspace{-3.5mm}

\end{figure*}

\begin{figure}[t]
    \centering
    \includegraphics[scale=0.31]{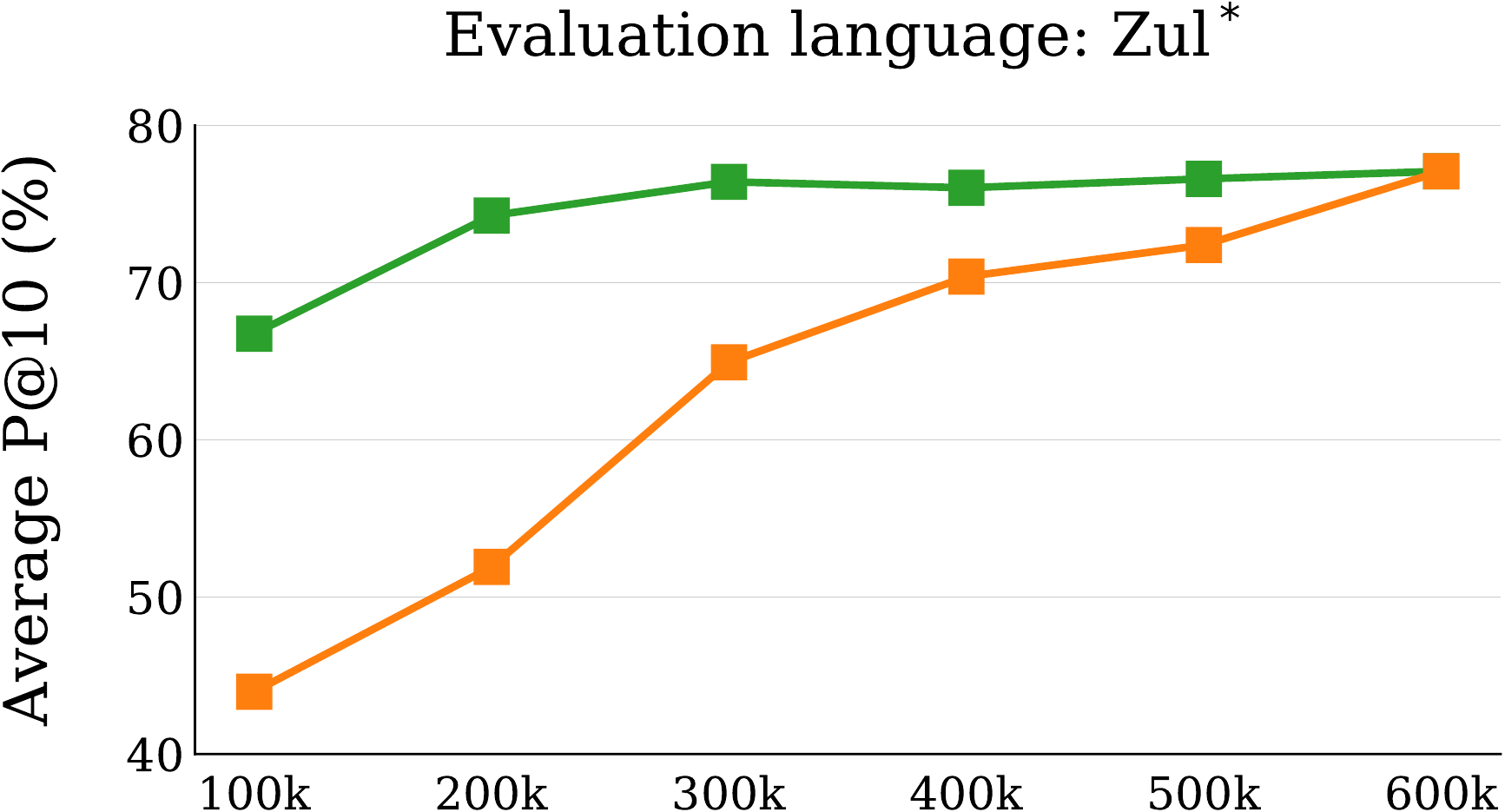}    
    \captionsep
    \caption{QbE results %
    on %
    Zul %
    using the same sequences of multilingual models as in Figure~\ref{fig:multi_increment}.}
    \label{fig:qbe_increment}
    \vspace{-4mm}
\end{figure}

\subsection{Multilingual evaluation}

The cross-lingual experiment above was in large part our inspiration for the subsequent analysis of multilingual models.
Concretely, we hypothesise that even better performance can be achieved by training on multiple languages from the same family as the target zero-resource language, and that this would be superior to multilingual models trained on unrelated languages.
Focusing on two evaluation languages, Zul and Sot from distinct language families, we
investigate this hypothesis
by training multilingual models with different language combinations, as shown in Table~\ref{tbl:multilingual}.  
Firstly, we see the best result on a %
language when all the training languages comes from the same family as the target.
Secondly, we see how the performance %
gradually decrease as the number of training languages related to the target drop.
Furthermore, notice the performance boost from %
including even just one training language related to the target compared to not including any.
E.g.\ on Sot we see a increase of more than 12\% absolute when adding just one related language (from 52.5\% and 51.9\% to 64.8\%). %

To further demonstrate the benefit of using training languages from the same family, %
we train a multilingual model for each evaluation %
language on all its related languages using a 10\% subset of the original data.
For both Zul and Sot, %
the subset models outperform the models where no related languages are used. %
E.g.\ on Zul, the Xho+Ssw+Nbl subset model outperforms the full Tsn+Nso+Eng model (no related languages) by more than 20\% in AP.
Moreover, this subset model (58.6\% in AP) even outperforms the Xho+Nso+Eng model (55.7\%) where all the training data from one related language are included and almost matches the AP when using two related languages (Xho+Ssw+Eng, 60.9\%).

These comparisons do more than just show the benefit of training on related languages: they also show that it is beneficial to train on a diverse set of related languages. %

\subsection{Adding more languages} %

In the above experiments, we controlled for the amount of data per language and saw that training on languages from the same family improves multilingual transfer.
But this raises a question:
will adding additional unrelated languages harm performance? %
To answer this, we systematically train two sequences of multilingual models on all six well-resourced languages, %
evaluating each target language as a new training language is added.

Same-different results for all five evaluation languages are shown in Figure~\ref{fig:multi_increment} and QbE results for Zul are shown in Figure~\ref{fig:qbe_increment}. As for Zul, the trends in the same-different and QbE results track each other closely for the other evaluation languages (so these are not shown here).
In the fist sequence of multilingual models (green), we start by adding the three Nguni languages (Xho, Ssw, Nbl), followed by the two Sotho-Tswana languages (Nso, Tsn), and lastly the Germanic training language (Eng).
The second sequence (orange) does not follow a systematic procedure.

On Zul, the green sequence, which starts with a related language (Xho), initially achieves a higher score compared to the orange sequence in both Figures~\ref{fig:multi_increment} and~\ref{fig:qbe_increment}.
Then, the score gradually increase by adding more related languages (Ssw, Nbl). %
Thereafter, adding additional unrelated languages (Nso, Tsn, Eng) show no performance increase.
In fact, AP decreases slightly after adding the two Sotho-Tswana languages (Nso, Tsn), but not significantly.
The orange sequence starts low on Zul until the first related language (Ssw) is added, causing a sudden  increase.
Adding the Germanic language (Eng) has little effect.
Adding the last two related languages (Nbl, Xho) again causes the score to increase.
A similar trend follows for Sot and Afr, where adding related languages causes a noticeable performance increase, especially when adding the first related language; after this, performance seem to plateau %
when adding more unrelated languages.
(Afr is the one exception, with a drop when adding the last language in the orange sequence).
On Tso, which does not have any languages from the same family in the training set, AP gradually increases in both sequences without any sudden jumps.
Although Ven isn't in the same family as Nso and Tsn at the lowest level of the tree in Figure~\ref{fig:hierarchy}, it belongs to the same family {(Sotho-Tswana)} at a higher level. %
This explains why it
closely tracks the Sot results.

Summarising these results, we see that adding unrelated languages generally does not decrease scores, but also does not provide a big benefit (except if it is one of the earlier languages in the training sequence, where data is still limited). In contrast, it seems that training on languages from the same family is again beneficial; this is especially the case for the first related language, irrespective of where it is added in the sequence.

	\section{Conclusion}
\label{sec:conclusion}

We investigated the effect of training language choice when applying a multilingual acoustic word embedding model to a zero-resource language.
Using word discrimination and query-by-example search tasks on languages spoken in Southern Africa, we
showed that training a multilingual model on languages related to the target is beneficial.
We observed gains in absolute scores, but also in data efficiency: you can achieve similar performance with much less data when training on multiple languages from the same family as the target.
We showed that even including just one related language %
already gives a large %
gain. %
From a practical perspective, these results indicate that one should prioritise collecting data from related languages (even in modest quantities) rather than collecting more extensive datasets from diverse unrelated families, when building multilingual acoustic word embedding models for a zero-resource language.

\vspace{2pt}
{\eightpt
\noindent \textbf{Acknowledgements.}
This work is supported by the South African NRF (120409), a Google Faculty Award, and support from the Stellenbosch University School of Data Science and Computational Thinking.
We thank Ewald van der Westhuizen for the NCHLT forced alignments.}

	\newpage
	\bibliography{mybib}
	
\end{document}